\documentclass[11pt]{article}

\usepackage[final]{acl}

\usepackage{times}
\usepackage{latexsym}
\usepackage{multirow}
\usepackage[T1]{fontenc}

\usepackage[utf8]{inputenc}

\usepackage{microtype}

\usepackage{inconsolata}

\usepackage{graphicx}

\usepackage{tabularx}
\usepackage{multirow}
\usepackage{enumitem}
\usepackage{amsmath}
\usepackage{amssymb}
\usepackage{booktabs}
\usepackage{tcolorbox}

\usepackage{ragged2e}

\usepackage[table]{xcolor}
\definecolor{em}{gray}{0.9}
\newcommand{\cem}{\cellcolor{em}}
\title{SEARCH-R: Structured Entity-Aware Retrieval with Chain-of-Reasoning Navigator for Multi-hop Question Answering}

\author{
 \textbf{Yuqing Fu\textsuperscript{1$^\dagger$}},
 \textbf{Yimin Deng\textsuperscript{1,2$^\dagger$}},
 \textbf{Wanyu Wang\textsuperscript{1}},
 \textbf{Yuhao Wang\textsuperscript{1}},
\\
 \textbf{Yejing Wang\textsuperscript{1}},
 \textbf{Hongshi Liu\textsuperscript{1}},
 \textbf{Yiqi Wang\textsuperscript{4}},
 \textbf{Xiao Han\textsuperscript{3}},
\\
 \textbf{Maolin Wang\textsuperscript{1}},
 \textbf{Guoshuai Zhao\textsuperscript{2}},
 \textbf{Yi Chang\textsuperscript{5}},
 \textbf{Xiangyu Zhao\textsuperscript{1*}}
\\
 \textsuperscript{1}City University of Hong Kong,
 \textsuperscript{2}Xi'an Jiaotong University,
 \textsuperscript{3}Zhejiang University of Technology,\\
 \textsuperscript{4}Michigan State University,
 \textsuperscript{5}Jilin University
\\
 \small{
   \textbf{Correspondence:} \href{mailto:yuqingfu5-c@my.cityu.edu.hk}{yuqingfu5-c@my.cityu.edu.hk}, \href{mailto:xianzhao@cityu.edu.hk}{xianzhao@cityu.edu.hk}
 }
}

\begin{document}
\maketitle
\renewcommand\thefootnote{\relax}
\footnotetext{$^\dagger$ Co-first authors with equal contribution.}
\footnotetext{* Corresponding author.}
\begin{abstract}

Multi-hop Question Answering (MHQA) aims to answer questions that require multi-step reasoning.~It presents two key challenges: generating correct reasoning paths in response to the complex user queries, and accurately retrieving essential knowledge in the face of potential limitations in large language models (LLMs).~Existing approaches primarily rely on prompt-based methods to generate reasoning paths, which are further combined with traditional sparse or dense retrieval to produce the final answer. However, the generation of reasoning paths commonly lacks effective control over the generative process, thus leading the reasoning astray. Meanwhile, the retrieval methods over-rely on knowledge matching or similarity scores rather than evaluating the practical utility of the information, resulting in retrieving homogeneous or non-useful information. Therefore, we propose a Structured Entity-Aware Retrieval with Chain-of-Reasoning Navigator framework named SEARCH-R. Specifically, SEARCH-R trains an end-to-end reasoning path navigator, which is able to provide a powerful sub-question decomposer by fine-tuning the Llama3.1-8B model. Moreover, a novel dependency tree-based retrieval is designed to evaluate the informational contribution of the document quantitatively. Extensive experiments on three challenging multi-hop datasets validate the effectiveness of the proposed framework.~The code and dataset are available at:~\url{https://github.com/Applied-Machine-Learning-Lab/ACL2026_SEARCH-R}.
\end{abstract}

\section{Introduction}

Question Answering (QA) focuses on finding appropriate answers for given queries and is a key task in natural language processing~\citep{ojokoh2018review,mishra2016survey}.
To enhance the performance of QA task, large language models (LLMs) have been employed and demonstrated effectiveness~\citep{siriwardhana2023improving,arslan2024survey}. Nevertheless, even if LLMs possess powerful reasoning capabilities, they present limitations on multi-hop question answering (MHQA) tasks, which involve complex queries such as: ``\textit{What is the name of the famous bridge located in the birthplace of the composer of Nulla in mundo pax sincera}?''. In such scenarios, accurately retrieving all relevant information and demanding LLMs to deduce implicit logical relationships remains challenging. In other words, the information retrieval and reasoning process requires in-depth collaboration, which is difficult to accomplish in one step.

To overcome this limitation, existing approaches employ multi-turn retrieval-augmented generation (RAG) to facilitate multi-step reasoning by LLMs through progressive information retrieval~\citep{trivedi2022interleaving,shao2023enhancing,jiang2025retrieve}. Specifically, it first generates a reasoning path or some sub-questions, and then retrieves the necessary external knowledge to answer them iteratively until a final answer is obtained. In this process, the quality of the reasoning path and retrieved external knowledge are particularly crucial, thus acting as two key components in MHQA tasks.

Recent research has advanced along two critical fronts. The first focuses on reasoning path generation, where existing works employ prompt-based methods, generating a chain-of-thought reasoning path or a series of simpler sub-questions by LLMs~\citep{wang2023plate}. For example, Iter-RetGen~\citep{shao2023enhancing} iteratively breaks down complex questions by leveraging outputs from previous steps. Meanwhile, other prompt-based methods dynamically generate sub-questions by assessing the query's complexity or identifying knowledge gaps~\citep{jeong2024adaptive,jiang2023active,liu2025large,xu2025harnessing}. In a parallel line of work,  non-prompt-based methods emerge to enhance the generation process. These approaches may leverage retrieved documents and annotations to train a dedicated generator or employ logic-driven filtering to distill high-quality training data from LLM outputs~\citep{zhuang2024efficientrag,ye2025optimizing,zhang2025llm}. However, these methods do not enable LLMs to acquire the ability to decomposing sub-questions: Prompt-based methods lack precise control over the sub-question generation process, while alternative approaches overemphasize logical optimality over answer accuracy on training data, which both lead to high stochasticity in reasoning planner and unstable QA performance. 


The second front focuses on information retrieval, where commonly used techniques include keyword matching~\citep{trivedi2022interleaving} such as BM25, and dense retrieval, which computes similarity scores for document retrieval~\citep{jeong2024adaptive,liu2025llmemb,wang2025rethinking}.~Improvements have also been made to existing methods~\citep{zhang2026search,zhang2025deep,wu2026deepresearch,jia2024mill}.~ChainRAG~\citep{zhu2025mitigating} constructs documents into a sentence graph, retrieves seed sentences through computing embedding similarity scores, then expands the associated information from the seed sentences. These methods mainly consider semantic similarities and the knowledge matching degree. However, such processes may overcome large amounts of homogeneous yet irrelevant information, which may not provide the necessary knowledge injection for LLMs.~Beyond traditional keyword-based or dense retrieval methods, tree-based retrieval offers inherently hierarchical constraints through its parent-child node relationships, which explicitly model interdependencies. This inspired our adoption of dependency parse trees, which structure a sentence into a tree that captures relational semantics among tokens. Tokens at different layers of this tree inherently correspond to varying levels of salience within the sentence, thereby enabling the derivation of a quantifiable informational representation.

Therefore, we present SEARCH-R, a three-stage MHQA framework that addresses these limitations by incorporating a robust reasoning path navigator, an entity-centric quantitative information retrieval module, and an iterative answer generator. Specifically, first, we construct a small set of high-quality data to train an end-to-end sub-question decomposer with only 8B parameters, which demonstrates strong generalization across multiple benchmarks. Second, we design a document entity informativeness metric that leverages dependency parsing to quantify the informational contribution of document entities, enhancing the accuracy of multi-document retrieval in MHQA tasks. Finally, guided by high-quality reasoning chains and supplemented with accurate external knowledge, the framework iteratively provides the final answer.

In summary, our contribution is threefold:

\begin{itemize}[leftmargin=*]
    \item We propose SEARCH-R, an efficient multi-hop question answering framework that enhances LLMs’ reasoning capabilities on complex questions through an optimized CoT architecture. 
    \item We design a multi-source information retrieval paradigm, present a new approach for entity-aware information quantification retrieval, providing LLMs with more precise and comprehensive evidence from multi-document contexts.
    \item We conduct extensive experiments on three challenging MHQA datasets, and the results demonstrate that our method achieves significant improvements over the best baseline.
\end{itemize}


\section{Method}

\subsection{Problem Formulation}
For a given complex question $q$ and a document set $\mathcal{D}$, the MHQA task aims to retrieve knowledge $\mathcal{K}\subset\mathcal{D}$ related to $q$ from $\mathcal{D}$, then generate an answer $a$ to the question $q$.

\subsection{Overall Framework}

\begin{figure*}[t]
\centering
    \includegraphics[width=\textwidth]{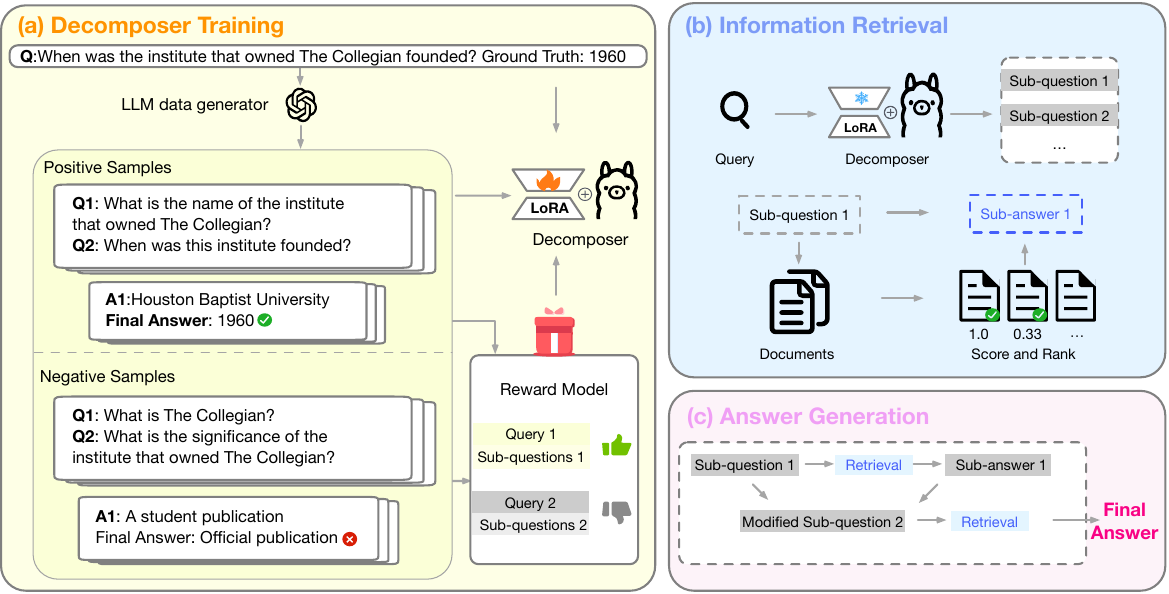}
  \caption{Overall structure of the proposed SEARCH-R framework.}
  \label{fig:overall}
\end{figure*}

Figure~\ref{fig:overall} shows the overall of our framework, which consists of three stages: \textbf{Reasoning Path Generation}, \textbf{Knowledge Retrieval}, and \textbf{Answer Generation}. First, we use LLMs to generate multiple reasoning paths for each question, then employ an end-to-end approach to filter out high-quality data and fine-tune a reasoning
path navigator.~Next, based on the reasoning paths, the process moves to the retrieval stage, where a combination of dense retrieval and entity informativeness is applied to retrieve the most relevant knowledge related to the question. Finally, multiple sub-questions and sub-answers are integrated to generate the final answer.

\subsection{Reasoning Path Generation}

Prompt-based approaches struggle to constrain the quality of reasoning paths generated by LLMs~\citep{shao2023enhancing,jeong2024adaptive,li2025towards,gao2025llm4rerank,fu2025unified}, and the latest improved methods rely on the logical accuracy of the training data rather than the correctness of the final answers\citep{zhuang2024efficientrag,ye2025optimizing}. To address these limitations, we propose and adopt an end-to-end training strategy to develop an efficient and highly generalizable reasoning path generator, which decomposes complex questions into a logically coherent sequence of simpler, answerable sub-questions.

To enable the model to learn how to generate reasoning paths for complex questions, we demand high-quality data. Considering the diversity of the training data, in the initial step, we leverage LLMs to break down each complex question $q$ into multiple varied sequences of sub-questions. Details of data Generator are shown in Appendix~\ref{data_generator} These reasoning paths are subsequently categorized by final answer consistency. The set of correct mappings is then employed for supervised fine-tuning (SFT) a 7B sub-question decomposer $M_{s}$:
\begin{align}
\mathcal{S}_{t},\mathcal{S}_{f}&=f_{c}(\operatorname{LLM}(q)) \\
M_{s}&=\operatorname{SFT}(\mathcal{S}_{t})
\end{align}
where $\mathcal{S}_{t},\mathcal{S}_{f}$ denote the two maps generated by LLMs classified by answer consistency, $\mathcal{S}_t$ represents question-subquestion pairs with a correct final answer, $\mathcal{S}_f$ represents the incorrect ones. $f_{c}$ denotes the classify operation.

To further enhance the model's decompose capability, we fine-tune the aforementioned SFT model using a reinforcement learning (RL) approach~\citep{zhao2019deep} continuously. This process contains two stages: \textbf{Reward Model Training} and \textbf{Policy Optimization} with Proximal Policy Optimization Algorithm ~\citep{schulman2017proximal}.

\textbf{Reward Model Training} To evaluate the quality of generated reasoning paths, we designed a reward model as a scoring function, where the target reward scores are computed based on F1 scores, which measures the matching level between the final answer $a_{f}$ and the ground truth $a_g$. Finally, we obtain a reward model $R_{\phi}$ which can score the quality of decomposed sub-questions.
\begin{equation}
    \label{loss_rm}
    \mathcal{L}_{\text{RM}} = \frac{1}{N} \sum_{i=1}^{N} \left( r_{p}^{(i)} - r_{t}^{(i)} \right)^2
\end{equation}
\begin{equation}
    r_{t} = \operatorname{F1}(a_{f}, a_{g})
\end{equation}

Equation~\eqref{loss_rm} shows the loss function $\mathcal{L}_{\text{RM}}$ of the reward model. $N$ represents the batch size, $r_{p}$ and $r_{t}$ denote the predicted and target reward score.

\textbf{PPO Fine-tuning} Afterward, leveraging the training signals provided by $R_{\phi}$, we train a policy network $\pi_{\theta}$ to shift the model's learning objective from imitating input and output data by SFT towards acquiring the capability to decompose complex questions by Proximal Policy Optimization (PPO). Specifically, the optimization goal of PPO is to maximize the expected reward while limiting the magnitude of policy updates. In our method, we optimize the policy network to generate high-quality sub-questions for an input $q$, where quality is measured by the score provided by the frozen $R\phi$. The loss function $\mathcal{L}_{\text{CLIP}}(\theta)$ of PPO is as follows:

\begin{equation}
    \mathcal{L}_{\text{CLIP}}(\theta) = \mathbb{E}_{\hat{t}, \hat{h}} \left[ \min\left( r_t(\theta) \hat{A}_t, c(\theta)\hat{A}_t \right) \right]
\end{equation}
     
\begin{equation}
    c(\theta)=\text{clip}\left(r_t(\theta),\ 1-\epsilon,\ 1+\epsilon\right)
\end{equation}

\begin{equation}
r_t(\theta) = \frac{\pi_\theta(a_t \mid s_t)}{\pi_{\theta_{\text{old}}}(a_t \mid s_t)}
\end{equation}

where $\theta$ denotes the trainable parameters of the policy model and $t$ represents the timestep. $c$ denotes the clip function. $r(\theta)$ is the probability ratio of new and old policy, $\pi_\theta(a\mid s)$ defines the probability of taking action $a$ in state $s$ under the current policy. $\epsilon$ is a hyperparameter, which prevents the network from updating too drastically by the clipping function.$ \hat{A}$ denotes the estimated advantage between action value $Q_t$ and state value $V(s)$. More discussion and mathematical demonstration of PPO can be found in the original paper.

Thus far, we have successfully trained an optimal answer-oriented reasoning path generator for complex questions, enabling the model to obtain the capability to decompose the sub-questions.

\subsection{Knowledge Retrieval}

\begin{figure}[t]
\centering
    \includegraphics[width=0.7\columnwidth]{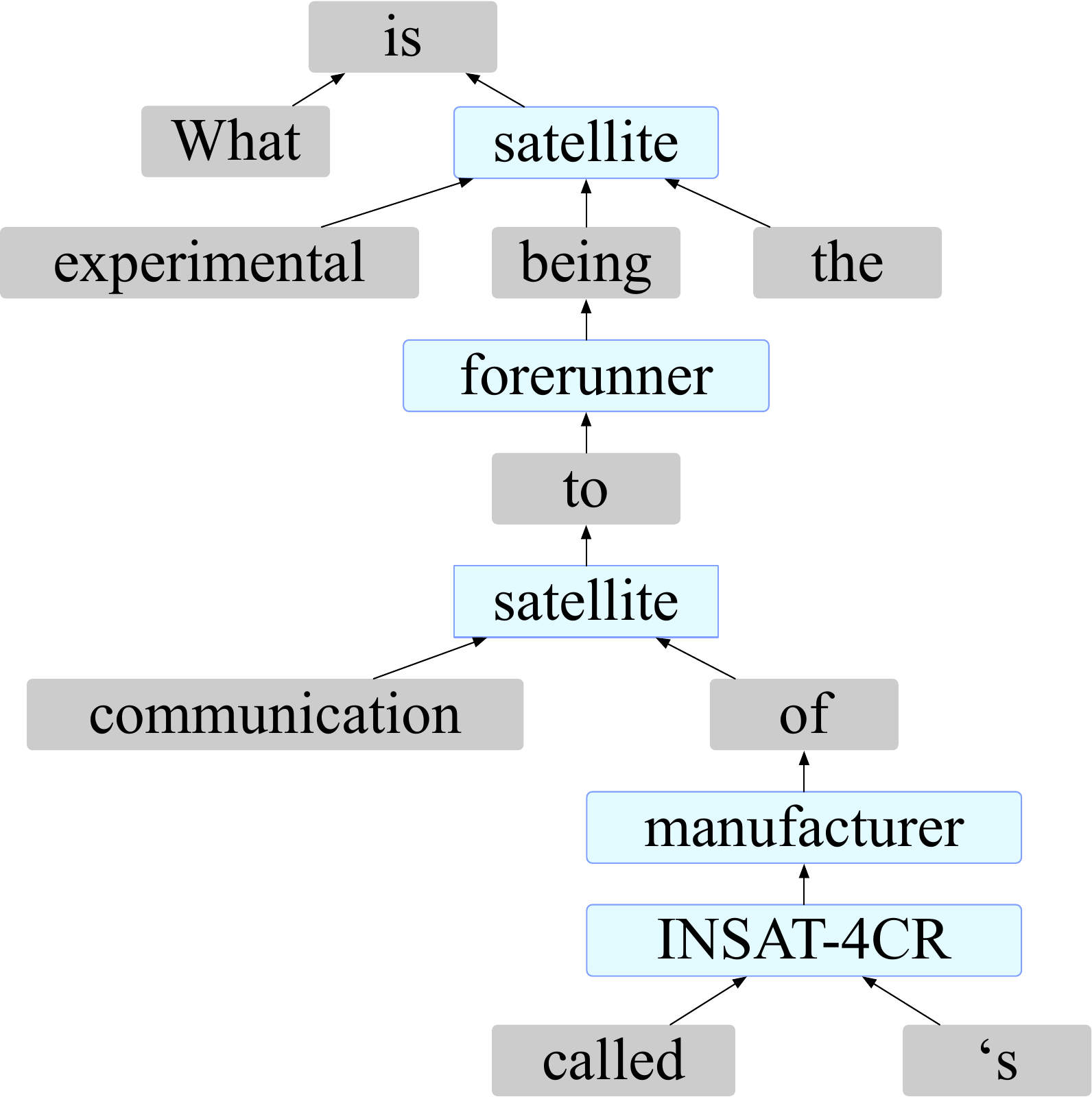}
  \caption{Structure of a dependency parsing tree. The original sentence is ``\textit{What is the experimental satellite being forerunner to communication satellite of INSAT-4CR’s manufacturer called?}'' The light-blue blocks represent the entity components.}
  \label{fig:tree}
\end{figure}

After obtaining the decomposed sub-questions, the process of retrieving relevant knowledge begins. In practice, although dense retrieval can capture certain semantic relationships within the information, it often struggles to identify truly useful content from items with high similarity scores due to issues such as information homogenization and the inherent limitations of embedding models in capturing fine-grained semantic distinctions~\citep{jeong2024adaptive,trivedi2022interleaving}. Therefore, instead of relying solely on traditional retrieval approaches, which compute similarity scores or perform keyword matching on knowledge entities, we assess the informational value of the documents by analyzing their syntactic structure, aiming to identify those segments that contribute most substantially.
We propose a dependency tree-based method to assess the informational contribution of document entities. First, we extract entities from the document using Named Entity Recognition. For each sentence, we build a dependency parse tree and compute the structural importance of each entity based on the maximum number of child nodes of its constituent words in the tree. The aggregate importance of an entity is defined as the sum of these $\operatorname{max\_child}$ counts across all sentences in the document. All entities are then ranked by their importance scores to form an ordered list $\mathcal{M}_r$. The final informativeness score $S_{doc}$ for a document is calculated as the sum of reciprocal ranks of all question entities in $\mathcal{M}_r$. This score is used to select the most informative document segments, which are combined with semantically similar passages retrieved via dense retrieval for subsequent processing. This retrieval process can be formulated as the mathematical expression below:
\begin{align}
\label{eq:dt}
    \operatorname{imp}(e_c)&=\sum_{s_c\in S_{doc}}\operatorname{max\_child}(w,\operatorname{DT}(s_c)),w\in e_c \\
    \mathcal{M}_r(e_c)&=\operatorname{rank}(\operatorname{imp}(e_c)),\forall e_c\in D \\
    s_{doc}&=\sum_{e_q\in Eq}\frac{1}{\mathcal{M}_r(e_q)}
\end{align}
where $s_c$ denotes a sentence from document $D$, and $e_c$ refers to entities extracted from $D$, $w$ represents the constituent words of $e_c$. The function $\operatorname{DT}$ builds the dependency parse tree for a sentence. Here, $\operatorname{max\_child}$ is defined as the maximum number of direct and indirect child nodes of word $w$ in the dependency tree of a sentence. 

Upon acquiring the document fragments, our retrieval employs a dual-path strategy, which combines the top-15 text chunks ranked by entity informativeness score and the top-10 selected by similarity score. We determined the number of similarity-based text chunks through separate experiments, top-10 chunks achieve optimal performance, as increasing or decreasing will reduce accuracy.

\subsection{Answer Generation}

After generating the reasoning path and retrieving relevant knowledge, answers are generated sequentially, question by question. Before answering a sub-question, the system first determines whether it is necessary to reformulate according to its current state. If the sub-question requires, the framework reformulates it by integrating the contextual history of previous sub-questions and their corresponding answers. This refined question is then used for subsequent retrieval and answer generation in the next step. This iterative process of ``\textit{answering-reformulating-retrieving-answering}'' continues until the entire reasoning chain is complete.




\section{Experiments}


\subsection{Experimental Settings}

\textbf{Datasets.} We evaluate the performance of our proposed method on three representative multi-hop benchmark datasets including MuSiQue, HotpotQA, and 2Wiki. We conducted our experiments using the identical data and processing procedures as ChainRAG~\citep{zhu2025mitigating}, rather than processing the raw dataset directly. Table~\ref{tab:musique} presents the statistics of three benchmarks.

MuSiQue~\citep{trivedi2022musique} is a challenging multi-hop dataset with 25k questions. These questions involve logical hops ranging from two to four. These questions feature a diverse range of chaining patterns. Specifically, except for structure of ``$A\to B\to C$'', MuSiQue also contains branching chain structure like ``$A\to C, B\to C, C\to D$''. 

HotpotQA~\citep{yang2018hotpotqa} is a multi-hop question answering dataset constructed from Wikipedia. It encompasses a diverse range of question types, including entities, time, numerical values, and comparisons. A key feature of this dataset is its provision of explanatory annotations, which facilitate the evaluation of reasoning processes.

2Wiki~\citep{ho2020constructing} is built upon structured and unstructured data of Wikipedia and presents a greater challenge than HotpotQA. 2Wiki provides evidential information for reasoning in the form of ``\textit{(subject entity, property, object entity)}'' triplets. The question contains four types: comparison, inference, compositional, and bridge comparison.

\textbf{Baselines and Metrics.} To evaluate the effectiveness of our approach, we choose several foundational or recent baselines for comparison: NativeRAG, Iter-RetGen, HippoRAG and ChainRAG. To ensure fairness, we configured NativeRAG's document retrieval to use the same similarity scoring model and chunking strategy as our method. Except for NativeRAG, we selected a set of recent MHQA methods for comparison: Search-R1~\citep{jin2025searchr1trainingllmsreason}, Iter-RetGen~\citep{shao2023enhancing}, ChainRAG~\citep{zhu2025mitigating}, and HippoRAG~\citep{jimenez2024hipporag} combined with IRCoT~\citep{trivedi2022interleaving}. We report the performance of Iter-RetGen, HippoRAG, and the result of ChainRAG (GPT-4o-mini) on MuSiQue from original report of ChainRAG. Three prevalent LLMs Llama3.1-8B~\citep{dubey2024llama}, DeepSeek-V3~\citep{liu2024deepseek} and GPT-4o-mini are conducted as our backbones. We attempted to incorporate GraphRAG~\citep{2024From} as an additional baseline, however, due to its suboptimal performance, it was ultimately excluded from the final comparison. The results of GraphRAG on MuSiQue is provided in Appendix~\ref{appendix:experiments}. The metrics we report in the main table are the F1 score and the exact match (EM) score. The baseline methods are further detailed in Appendix~\ref{baselin_details}. 

\begin{table}[t]
  \centering
  
  \begin{tabular}{lcccc} 
    \hline  
     \textbf{Benchmarks}& \textbf{Train} & \textbf{Val} & \textbf{Test}  \\
    \hline  
    MuSiQue & 19,938  & 2,417   &2,459     \\
    HotpotQA & 90,564 & 7,405& 14,810\\
    2WikiMQA &167,454 & 12,576 & 12,576 \\
    
    \hline  
  \end{tabular}
  \caption{Statistics of our three benchmarks. Train, Val, and Test represent the number of samples in training set, validation set and test set, respectively.}  
  \label{tab:musique}  
\end{table}

\begin{table*}[ht!]
\centering
\begin{tabular}{l c c c c c c c}
\toprule
\multirow{2}{*}{\textbf{Models}}&\multirow{2}{*}{\textbf{Backbone}}&\multicolumn{2}{c}{\textbf{MuSiQue}} & \multicolumn{2}{c}{\textbf{HotpotQA}} & \multicolumn{2}{c}{\textbf{2Wiki}} \\
\cline{3-8}
 &  & F1 & EM & F1 & EM & F1 & EM\\
\hline
NativeRAG&\multirow{4}{*}{Llama-3.1-8B} & 23.82&15.50 & 47.00 & 33.50 & 41.13 &32.00\\
Search-R1& &24.67 & 18.50& 43.52 & 30.50 & 37.31 &32.00\\
ChainRAG& & 36.02 & 27.00 & 45.21 & 34.00 & 33.44 & 25.00 \\
\cem{SEARCH-R$^*$} &\cem{}& \cem{\textbf{43.56}}	&\cem{\textbf{34.00}}	&\cem{\textbf{49.09}}&	\cem{\textbf{36.00}}&	\cem{\textbf{41.65}}&	\cem{\textbf{34.00}}\\
\hline
NativeRAG&\multirow{2}{*}{DeepSeek-V3} & 24.23 & 15.50 & 46.19 & 32.50 & 40.04 & 31.00\\
ChainRAG& & 39.05 & 27.50 & 53.99 & 41.00 & 61.40 & 52.00\\
\cem{SEARCH-R$^*$}&\cem{Llama-3.1-8B+DeepSeek-V3 }&\cem{\textbf{53.36}} & \cem{\textbf{44.50}} & \cem{\textbf{61.63}} &\cem{\textbf{46.00}} & \cem{\textbf{62.49}} & \cem{51.50}\\
\hline
NaiveRAG &\multirow{4}{*}{GPT-4o-mini} & 29.82 & 19.00 & 56.92 & 42.00 & 50.61 & 42.50 \\
Iter-RetGen& & 38.41 & 33.00 & 57.77 & 42.00 & 58.43 & 50.50 \\
HippoRAG&  & 46.50 & 28.50 & 56.12 & 40.00 & 62.38 & 48.00 \\
ChainRAG& & 50.54 &37.00 & 59.13 & 44.50 & 56.61 & 48.00\\
\cem{SEARCH-R$^*$} &\cem{Llama-3.1-8B+GPT-4o-mini} & \cem{\textbf{55.68}} &	\cem{\textbf{45.00}}&	\cem{\textbf{60.06}}	&\cem{44.00}	&\cem{\textbf{63.86}}	&\cem{\textbf{52.50}}\\
\bottomrule
\end{tabular}
\\[1ex]
\caption{Performance comparison of different models (in percentage) on three datasets. $^*$ indicates the statistically significant improvements (two-sided t-test with $p<0.05$) over the best baseline.}
\label{table:main}
\end{table*}

\begin{table}[t!]
\setlength\tabcolsep{9.pt}
\centering\begin{tabular}{lcccc}
\toprule
\multirow{2}{*}{\textbf{Model}} & \multicolumn{2}{c}{\textbf{GPT-4o-mini}} & \multicolumn{2}{c}{\textbf{Llama-3.1-8B}}  \\
\cline{2-5}
& F1 & EM & F1 & EM \\
\hline
Ours& \textbf{55.68} & \textbf{45.00} & \textbf{43.56} &\textbf{34.00}  \\
w/o Dec. & 53.17  & 40.50 & 35.15 & 25.50 \\
w/o QIR & 48.78 & 39.00 & 35.37 & 29.00 \\
w/o PPO & 53.03 & 41.00 &36.98 & 27.50\\
\bottomrule
\end{tabular}
\caption{Results of ablation study (in percentage). ``w/o'' annotates to remove the corresponding module. ``Dec.'', ``QIR'' represents the decomposer and quantitative information retrieval module. We conduct the ablation study on MuSiQue dataset.}
\label{table:ablation_study}
\end{table}

\textbf{Implement Details.} Our experiments are implemented on 4 Intel(R) Xeon(R) Gold 6348 @2.60GHz CPUs and 4 NVIDIA A800-80GB(80GB) GPUs, with optimizer AdamW. We first generated two distinct reasoning paths for each of the 2,000 complex questions. By applying answer consistency filtering, we selected 300 high-quality ``\textit{question–subquestion}'' sets that led to correct answers, which were used to train the fine-tuned model. The learning rate of fine-tuning stage is 2e-5. All generated data was then split, half was used to train a reward model, and the other half to train the PPO policy model. The training of both the fine-tuned model and the reward model was completed in less than one hour, while training the PPO network required approximately three hours. Each experiment was replicated over five times. We use en\_core\_web\_trf for syntactic dependency parsing, sentence-transformers/all-MiniLM-L6-v2 for similarity computing, and NER was performed using the model of flair/ner-english-large. The top $k$ hyperparameter for the document entity informativeness score is 15. More details of data collection are provided in Appendix~\ref{data_generator}.

\subsection{Main Results}

The main results are presented in Table~\ref{table:main}.
Specifically, when trained solely on the MuSiQue, the F1 score of SEARCH-R on MuSiQue outperforms the best baseline by 20.9\% on Llama3.1-8B, 36.6\% on DeepSeek-V3, and 10.7\% on GPT-4o-mini. This indicates the superior generalization capability of the sub-question decomposer across all three benchmark datasets. Besides, the overall performance shows improvement when a more powerful backbone is employed for the final answer generation, which improves 22.5\% when replacing Llama3.1-8B by Deepseek-V3 and 27.8\% by GPT-4o-mini. Moreover, our method with an 8B-perameter-decomposer performs better than methods with powerful backbones in the whole process. 

These results demonstrate two key findings: First, the decomposition capability can be learned efficiently by a smaller model. Second, answer generation is highly dependent on the LLMs' capacity for knowledge integration and coverage. We also evaluate the precision and recall metrics, and the results are reported in Appendix~\ref{appendix:experiments}.

\subsection{Ablation Study}

To evaluate the contribution of each component, we conducted series of ablation studies, the results of which are shown in Table~\ref{table:ablation_study}. The two sets of columns represent results obtained using different backbone models for answer generation.

\textbf{Ablating Sub-question Decomposition.} To evaluate the gain of our trained sub-question decomposer, we replaced it with LLMs without task-specific training, using the identical prompt from our training data generation stage to decompose original questions. 
The results show that the sub-question decomposer leads to a 23.9\% relative gain on F1 score when using the Llama as the answer generator. Moreover, it even performs 4.7\% relative gain on F1 score on GPT-4o-mini, which is a more powerful backbone. 
This indicates that our approach has enabled a compact 8B-parameter model to effectively master the skill of question decomposition, yielding a solution that is both highly efficient and exhibits strong generalization. 

\textbf{Ablating Quantitative Information Retrieval.}  We ablated the module for calculating quantified document informativeness and relied solely on traditional dense retrieval~(sentence-transformers/all-MiniLM-L6-v2 for similarity computing) to obtain relevant knowledge. The results demonstrate that incorporating document informativeness substantially enhances the accuracy of knowledge retrieval, thereby providing more reliable knowledge support for addressing sub-questions.

\textbf{Ablating PPO Fine-tuning.} To evaluate the contribution of PPO fine-tuning, we replace the PPO-tuned model with the SFT model as the sub-question decomposer. Results indicate that the model without PPO fine-tuning underperforms its PPO-tuned counterpart. This suggests that the SFT model primarily engages in imitation, learning input-output patterns from the data. In contrast, the PPO model, guided by scores from the reward model that assess decomposition quality, learns to optimize across multiple dimensions. These include the logic of decomposition, the interdependencies between sub-questions, and the knowledge coverage of the main question, thereby significantly enhancing its capability to generate effective reasoning paths for complex questions.

\begin{table}[t!]
\setlength\tabcolsep{10.pt}
\centering\begin{tabular}{lcc}
\toprule
\textbf{Models}& \textbf{F1} & \textbf{EM}\\
\hline
Ours& \textbf{43.56} & \textbf{34.00} \\
w/o PPO & 36.98 & 27.50 \\
w/o PPO w/SFT & 37.67 & 28.50\\
\bottomrule
\end{tabular}
\caption{Results of further analysis of w/o PPO. We conduct the study on MuSiQue by llama-3.1-8b for answering.}
\vspace{-3mm}
\label{table:ablation_ppo}
\end{table}
\begin{table}[t!]
\setlength\tabcolsep{10.pt}
\centering\begin{tabular}{lcc}
\toprule
\textbf{Decomposer}&  \textbf{F1} & \textbf{EM}\\
\hline
GPT-4o &  58.26 & 48.00 \\
Llama-3.1-8B(Ours) & \textbf{58.68} & \textbf{48.50} \\

\bottomrule
\end{tabular}
\caption{Results of GPT-4o on MuSiQue. Both of the answer generators are GPT-4o.}
\vspace{-3mm}
\label{table:gpt-4o}
\end{table}

To establish a fair comparison with the PPO approach, we supplemented the ablation study with an SFT-tuned Llama-3.1‑8B model trained on the same volume of data used for PPO, denoted as w/o PPO w/SFT. The results are summarized in Table~\ref{table:ablation_ppo}, which show that while increasing the SFT data (w/o PPO w/SFT) yields a slight gain over the small-data SFT baseline (w/o PPO), the improvement remains marginal. In contrast, our full PPO-based model delivers a substantially higher performance, underscoring that the gains from PPO cannot be attributed merely to data scaling of SFT. This confirms the distinct value of PPO in enhancing model capability beyond what is achievable through supervised fine-tuning alone.

Furthermore, we provide the results on GPT-4o, which is a more powerful backbone, the results are in Table~\ref{table:gpt-4o}. As evidenced by the results, when the sub-question decomposer is replaced with the more powerful GPT-4o, our method, utilizing a model with only 8 billion parameters, still slightly outperforms the prompt-based GPT-4o. This indicates that the task of sub-question decomposition does not rely heavily on the internal knowledge of larger, more capable LLMs. It is possible to train a model with fewer parameters to acquire decomposition capability, achieving performance close to or even surpassing that of a much larger model.

\begin{table}[t]
    \centering
    
    \begin{tabular}{p{0.9\columnwidth}}
        \toprule
        \textbf{Case Study} \\
        \hline
        \textbf{Question}:~\textit{Who plays the wife of the producer of Here Comes the Boom in Grown Ups?}\\
        \textbf{Ground Truth}: \textit{Maria Bello}\\
        \textbf{Step 1: Sub-question Decompose} \\
        Sub-question 1: \textit{Who is the producer of Here Comes the Boom?}\\
        Sub-question 2: \textit{Who plays the wife of this producer in Grown Ups?}\\
        \textbf{Step 2: Retrieval and Answer} \\
        Retrieval and Answer for Sub-question 1: \textit{Kevin James}\\
        If Sub-question 2 should be rewritted? \textit{Yes}\\
        Rewritted Sub-question 2: \textit{Who plays the wife of Kevin James in Grown Ups?}\\
        Retrieval and Answer for Sub-question 2: \textit{Maria Bello}\\
        \textbf{Step 3: Intergrating for final answer }
        Final Answer: \textit{Maria Bello}\\
        \bottomrule
    \end{tabular}
    \caption{Case Study of Our Framework.}
    \label{tab:case_study}
    \vspace{-3mm}
\end{table}

\subsection{Additional Analysis}

\textbf{Case Study} Table~\ref{tab:case_study} illustrates a complete end-to-end reasoning process for the complex query: ``\textit{Who plays the wife of the producer of
Here Comes the Boom in Grown Ups?}''. Our trained sub-question decomposer first breaks it down into simpler sub-questions: ``\textit{Who is the producer of Here
Comes the Boom?}'',``\textit{Who plays the wife of this producer in Grown Ups?}''. Then the process begins by retrieving necessary knowledge for the initial sub-question, yielding the answer ``\textit{Kevin
James}''. Subsequent sub-questions are then iteratively rewritten and answered. For instance, based on Sub-question 1 and its answer, the missing entity ``\textit{this producer}'' in Sub-question 2 is resolved, leading to its reformulation as``\textit{Who plays the wife of Kevin James in Grown Ups?}''. This refined question is then retrieved and answered, resulting in ``\textit{Maria Bello}''. Finally, all sub-questions and their answers are synthesized to produce the final answer.

\begin{figure}[t]
\centering
    \includegraphics[width=\columnwidth]{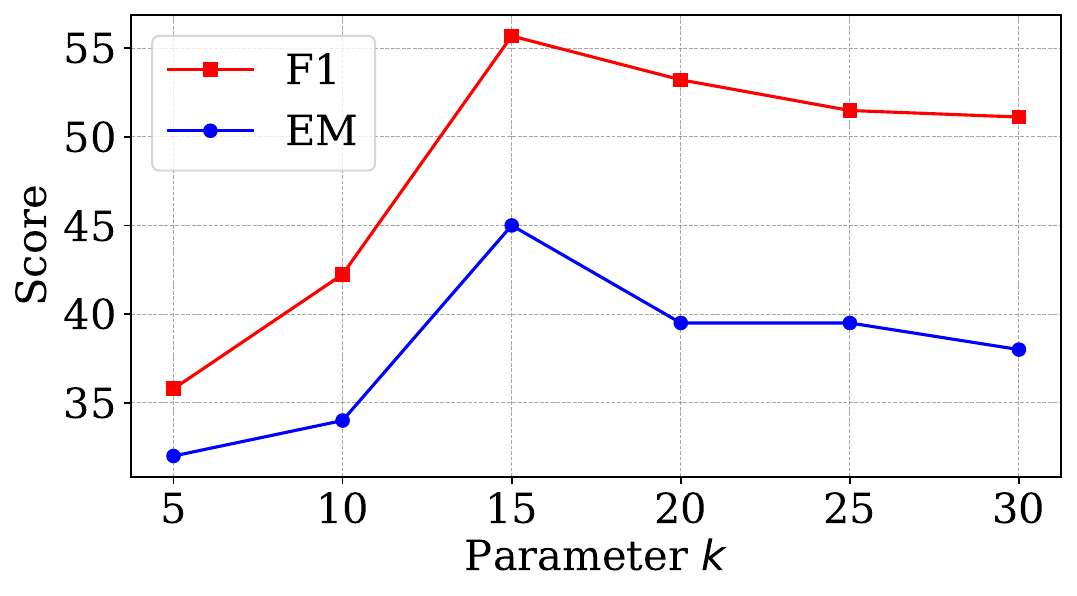}
  \caption{Impact of parameter $k$ on MuSiQue for Top-k Most Informative Documents.}
  \label{fig:topk}
\end{figure}

\textbf{Hyperparameter Study} Figure~\ref{fig:topk} illustrates the impact of the top $k$ parameter on answer accuracy. $k$ is the number of documents selected based on the informativeness score. As shown, accuracy is low at $k=5$ due to insufficient information coverage for answering the questions. When $k$ exceeds 15, accuracy plateaus and declines from its peak, which is attributable to the noise introduced by an excessive number of documents, overwhelming the LLM's ability to process the information effectively. The peak accuracy, 55.68\% of the F1 score and 45\% of the EM score are both achieved at $k=15$, representing an optimal balance where the retrieved documents provide adequate knowledge coverage without introducing extra noise.

\textbf{QIR Further Analysis} Furthermore, we introduced an additional experimental setup that removes the traditional dense retrieval component, relying solely on our QIR module(w/o dense retrieval). The results (Table~\ref{table:ablation_qir}) demonstrate that QIR provides complementary information not fully captured by dense retrieval alone. The combined use of both methods yields a synergistic effect, leading to higher answer accuracy.

For error analysis, we randomly mask 1 entity of each sub-question, retaining rest entities for informativeness calculation to simulate entity recognition failures. Results demonstrate that our method still outperforms similarity-only retrieval on F1 score, indicating that our entity-aware approach continues to provide useful information even when entity recognition is partially unsuccessful. 

\begin{table}[t!]
\setlength\tabcolsep{10.pt}
\centering\begin{tabular}{lcc}
\toprule
\textbf{Models}& \textbf{F1} & \textbf{EM}\\
\hline
Ours& \textbf{55.69} & \textbf{45.00} \\
w/o dense retrieval in QIR & 52.71 & 39.00 \\
masked QIR & 51.11 & 38.00\\
w/o QIR & 48.78 & 39.00 \\
\bottomrule
\end{tabular}
\caption{Results of further analysis of QIR. We conduct the study on MuSiQue by gpt-4o-mini for answering.}
\vspace{-3mm}
\label{table:ablation_qir}
\end{table}

\section{Related Works}

\subsection{LLMs and RAG}

In 2022, the hit of ChatGPT has aroused widespread attention to LLMs. Subsequently, numerous LLMs have demonstrated remarkable natural language processing capabilities. However, due to limitations of training data, the internal knowledge of LLMs is constrained, and the probabilistic generation mechanisms lead LLMs to produce hallucinations.
To mitigate this phenomenon, RAG~\citep{lewis2020retrieval} augments LLMs by integrating retrieved external knowledge into LLM's context, which has become a prominent method and applied extensively across diverse domains, including QA tasks. For example, RECOMP~\citep{xu2024recomp} condenses the retrieved documents into summaries and leverages them as contextual information; GraphRAG~\citep{edge2024local} constructs a knowledge graph from the documents, then retrieves relevant information by querying this graph for specific entities and subsequently generates a response. Since most questions in these methods are simple, the retrieval and generation can be accomplished within a single RAG round.

\subsection{Iterative RAG}

Single-step RAG, however, struggles with complex multi-step questions. Iterative RAG addresses this by dynamically generating sub-questions and retrieving relevant information through repeated cycles~\citep{jiang2025retrieve,xiong2024improving,shao2023enhancing}. Key challenges lie in determining the reasoning path and retrieval targets dynamically as the question state evolves.


On one hand, for reasoning path generation, IRCoT~\citep{trivedi2022interleaving} augments the chain-of-thought generation with the retrieved content, thereby enhancing its quality through improved factual consistency. Adaptive-RAG~\citep{jeong2024adaptive} dynamically adjusts its RAG iteration strategy based on the complexity of the question. FLARE~\citep{jiang2023active} first employs the LMs to identify knowledge gaps, then generates a logically coherent sequence of sub-questions that can be solved independently. While these methods for generating reasoning paths suffer from a lack of strict control during the generation process, leading to reasoning trajectories that can randomly deviate from the original query's intent. In contrast, our work employs an end-to-end methodology that focuses on enabling the model to grasp the principled decomposition of complex tasks.

On the other hand, information retrieval methods are broadly categorized into two paradigms: sparse retrieval and dense retrieval. Widely-used sparse methods like TF-IDF and BM25 rely on term frequency and document statistics to score documents, but often fail to capture latent semantic relationships. Dense retrieval addresses this by using text similarity models to capture contextual connections, though it remains constrained by data and model limitations, often retrieving homogeneous or irrelevant results. To overcome these issues, we integrate dependency parse trees to quantify document informativeness and combine them with dense retrieval. This hybrid approach captures hidden dependencies while assessing informational value, significantly improving retrieval efficiency.


\section{Conclusion}

In this paper, we propose SEARCH-R, a structured entity-aware retrieval framework 
for multi-hop question answering (MHQA). Our framework systematically addresses two core challenges: generating high-quality reasoning paths and retrieving precisely useful knowledge. Specifically, we propose an end-to-end reasoning path navigator 
to decompose complex questions into logically coherent sub-questions with strong generalization. 
Moreover, we design a quantitative document retrieval method based on dependency syntactic trees, which effectively evaluates the informational contribution of document entities beyond semantic similarity. 

\section*{Limitations}

Despite its strong performance, our framework has certain limitations. First, while our retrieval method improves relevance assessment via syntactic informativeness, it still relies on an external parser and entity recognition tools, which may introduce pipeline errors and domain dependency. Second, although the reasoning navigator generalizes well across datasets, its performance remains partly constrained by the quality and diversity of the initial LLM-generated training data. Future work may explore joint training of the retrieval and reasoning modules, and more efficient mechanisms for dynamic path planning and document scoring.

\subsubsection*{Acknowledgments}
This research was partially supported by National Natural Science Foundation of China (No.62502404), Hong Kong Research Grants Council (Research Impact Fund No.R1015-23, Collaborative Research Fund No.C1043-24GF, General Research Fund No. 11218325), Institute of Digital Medicine of City University of Hong Kong (No.9229503), Huawei (Huawei Innovation Research Program), Tencent (Tencent Rhino-Bird Focused Research Program, Tencent University Cooperation Project), Didi (CCF-Didi Gaia Scholars Research Fund), Kuaishou (CCF-Kuaishou Large Model Explorer Fund No. 2025008, Kuaishou University Cooperation Project), and Bytedance.

\bibliography{main}

\appendix

\section{Appendix}
\subsection{Data Collection and Generator}
\label{data_generator}

Our data collection process commence with a random sample of 2,000 instances from the MuSiQue training set. For each instance, we generate two reasoning paths(sub-questions) by gpt-4o-mini. Each sub-question is then processed by our retrieval module to gather information and is answered iteratively until a final answer is reached. Through this end-to-end process, we identify the reasoning paths that led to the correct answers, designating them as high-quality reasoning data. From this refined pool, we randomly select 300 question-reasoning path pairs to form the dataset for fine-tuning our SFT model. We will incorporate a detailed explanation of this data curation process.

The following prompt is used in the data generation phase, which leverages LLMs to produce two different raw reasoning paths per input, thereby creating diverse training data for the next step.

\begin{tcolorbox}[colframe=brown,
        width=1\linewidth,
        arc=1mm, 
        auto outer arc,
        title={Prompt Data Generator}]
        You are a question-decomposition AI assistant. Your sole mission is to split complex questions into the smallest set of essential sub-questions required for a complete answer. For each question, generate 2 distinct sub-question lists simultaneously.  \\
       Question: \underline{<QUESTION>}.\\
       Break down this question into minimal necessary sub-questions:\\
       ...
\end{tcolorbox}

The full version is provided in Table~\ref{tab:decomposer}.

\subsection{Baselines}
\label{baselin_details}
We compare our experiment with traditional baseline NativeRAG and the latest MHQA methods: Iter-RetGen, HippoRAG, and ChainRAG. Here are some details of our baselines.

\textbf{Search-R1} Search-R1 is an agentic approach based on reinforcement learning. We reproduce and train it using PPO on an amount of data equivalent in scale to that used for our method.

\textbf{Iter-RetGen} Iter-RetGen is an iterative RAG framework. Through multiple ``\textit{retrieval-generation}'' cycles, it leverages the output of each generation to guide the subsequent retrieval step, progressively refining the final answer.

\textbf{HippoRAG} HippoRAG is a graph-based retrieval method that constructs a Knowledge Graph from open domain information and retrieves answers by extracting named entities from the query.

\textbf{ChainRAG} ChainRAG is a graph-based question-answering framework that structures the entire information space into a sentence-level graph. It then retrieves relevant context by initiating searches from seed sentences, which are identified based on the user's query.

\subsection{Experiments Results}
\label{appendix:experiments}

Following the official GraphRAG~\citep{2024From} documentation, we constructed a knowledge graph and built the index on our dataset MuSiQue to reproduce the method. We use gpt-4o-mini as the backbone and the results are presented. 
\begin{table}[t!]
\setlength\tabcolsep{10.pt}
\centering\begin{tabular}{lcc}
\toprule
\textbf{Models}& \textbf{F1} & \textbf{EM}\\
\hline
GraphRAG& 21.4 & 6.00 \\
NativeRAG & 29.82 & 19.00 \\
Ours & \textbf{55.68} & \textbf{45.00}\\
\bottomrule
\end{tabular}
\caption{Results of GraphRAG, NativeRAG and SEARCH-R on MuSiQue by gpr-4o-mini for answering.}
\vspace{-3mm}
\label{table:graphrag}
\end{table}

As shown in \ref{table:graphrag}, GraphRAG performs poorly in this task, and similar obsevations have been reported in StepChain GraphRAG~\citep{ni2025}. This is primarily due to inconsistent quality in entity/relationship extraction during knowledge graph construction, leading to incomplete extraction of entities and relationships. As a result, the graph structure lacks critical nodes/edges, which ultimately causes retrieval failures. In contrast, NativeRAG relies solely on embedding-based similarity computation without introducing additional noise through complex technical pipelines. Our approach enhances the dense retrieval mechanism by incorporating an entity informativeness calculation module, which improves retrieval quality. Moreover, even when entity recognition fails, the dense retrieval component can still provide relevant information, thereby balancing retrieval accuracy and methodological robustness.

\begin{table}[t!]
\centering
\begin{tabular}{l  l c c}
\toprule
& &ChainRAG&  Ours\\
\hline
\multirow{2}{*}{\textbf{MuSiQue}}& Prec.& 39.65 &55.79  \\
& Recall & 38.85&57.59\\
\hline
\multirow{2}{*}{\textbf{HotpotQA}}&Prec. &62.78 &62.85  \\
& Recall&59.60&63.58\\
\hline
\multirow{2}{*}{\textbf{2Wiki}}&Prec.&57.25 & 62.18  \\
& Recall &57.84 & 68.66\\
\bottomrule
\end{tabular}
\\[1ex]
\caption{Precision and recall metrix of SEARCH-R (in percentage) on MuSiQue. SEARCH-R indicates the statistically significant improvements (two-sided t-test with $p<0.05$) over the best baseline.}
\label{table:extra_matrix}
\end{table}

Table~\ref{table:extra_matrix} presents a comparison of precision and recall between our method and our strongest baseline ChainRAG, both utilizing the GPT-4o-mini as backbone. The results demonstrate that our method outperforms the baseline across all three datasets in both precision and recall. Since the results of MuSiQue in Table~\ref{table:main} we report are from the original paper of ChainRAG, the corresponding precision and recall values are not available. Therefore, we report the results from our own reproduction of the experiment here. The F1 and EM scores of the reproduction results are 38.64 and 30.00.

\subsection{Time Complexity Analysis} 

Table~\ref{table:time_cost} shows the results on average reasoning time per question, which demonstrate a clear advantage of our method over other baselines. NativeRAG requires the least time, as it involves no optimization steps and the fewest LLM calls. ChainRAG shows a dramatic difference in time consumption with and without caching, due to its need to build a sentence graph for each question as a preprocessing step, which is a computationally expensive step. In contrast, our approach leverages dependency parse trees to rapidly estimate document informativeness, enabling both accuracy and highly efficient knowledge retrieval.

\begin{table}[t!]
\setlength\tabcolsep{12.pt}
\centering\begin{tabular}{lc}
\toprule
\textbf{Models}& \textbf{Time}\\
\hline
NativeRAG& 5.4s \\
ChainRAG (w/ cache)& 12.9s\\
ChainRAG (w/o cache)& > 1 min\\
Ours& 18.8s\\
\bottomrule
\end{tabular}
\caption{Average time required for reasoning per question across different methods under gpt-4o-mini.}
\vspace{-3mm}
\label{table:time_cost}
\end{table}

\begin{table*}[h!]
    \centering
    \newcolumntype{C}{>{\fontfamily{zi4}\selectfont}X}
    \begin{tabularx}{\textwidth}{C}
        \toprule
  You are a question-decomposition AI assistant. Your sole mission is to split complex questions into the smallest set of essential sub-questions required for a complete answer. For each question, generate 2 distinct sub-question lists simultaneously.\\\\
  \textbf{Core Principles to Follow Rigorously}: \\\\
1.Decompose Only When Necessary: Never split a question if it can be answered with a single piece of information. Only break it down when multiple distinct facts must be collected and connected to form the answer. \\
2.Zero Redundancy: Ensure no two sub-questions overlap in content or intent. Every sub-question must serve a unique purpose in answering the original query.\\
3.Single-Focus Sub-Questions: Each sub-question must target one and only one specific, non-negotiable piece of information.~Avoid vague or multi-topic sub-questions at all costs.\\
4.Minimalist Sub-Question Count: Keep the number of sub-questions as small as possible, aim for 2 sub-questions at maximum for any original query.\\
5.Output Restriction: Your response must consist only of a valid JSON array of sub-questions. Under no circumstances include explanations, notes, or extra text.\\\\
Strict Output Format Example:\\
Question: "Who plays the wife of the producer of Here Comes the Boom in Grown Ups?"\\
Your Output:\\
{["Who is the producer of Here Comes the Boom?", "Who plays the wife of this producer in Grown Ups?"]}\\\\
Question:\{Question\}\\
Break down this question into minimal necessary sub-questions:\\
        \bottomrule
    \end{tabularx}
    \caption{Prompt of sub-question decomposer.}
    \label{tab:decomposer}
\end{table*}

\label{sec:appendix}

\end{document}